\documentclass[letterpaper, 10 pt, conference]{ieeeconf}  

\IEEEoverridecommandlockouts                              
\usepackage{graphics} 
\usepackage{epsfig} 
\usepackage{mathptmx} 
\usepackage{times} 
\usepackage{amsmath} 
\usepackage{amssymb}  
\usepackage{upquote}
\usepackage{bbm}
\usepackage{scrlayer-scrpage}  
\title{\LARGE \bf
Assessing Alcohol Use Disorder: Insights from Lifestyle, Background, \\ and Family History with Machine Learning Techniques
}

\author{Chenlan Wang$^{1}$, Gaojian Huang$^{2}$, and Yue Luo$^{2}$
\thanks{$^{1}$ University of Michigan, Ann Arbor, USA}%
\thanks{$^{2}$ San José State University, San Jose, USA}%
}

\begin{document}

\maketitle
\thispagestyle{empty}
\pagestyle{empty}

\begin{abstract}
This study explored how lifestyle, personal background, and family history contribute to the risk of developing Alcohol Use Disorder (AUD). Survey data from the All of Us Program was utilized to extract information on AUD status, lifestyle, personal background, and family history for 6,016 participants. Key determinants of AUD were identified using decision trees including annual income, recreational drug use, length of residence, sex/gender, marital status, education level, and family history of AUD. Data visualization and Chi-Square Tests of Independence were then used to assess associations between identified factors and AUD. Afterwards, machine learning techniques including decision trees, random forests, and Naive Bayes were applied to predict an individual’s likelihood of developing AUD. 
Random forests were found to achieve the highest accuracy (82\%), compared to Decision Trees and Naive Bayes. Findings from this study can offer insights that help parents, healthcare professionals, and educators develop strategies to reduce AUD risk, enabling early intervention and targeted prevention efforts.
\end{abstract}


\section{INTRODUCTION}

In 2023, around 85\% of U.S. adults aged 18 and over reported alcohol consumption \cite{niaaa1}. 
Despite widespread awareness of alcohol's negative effects—ranging from short-term consequences like hangovers and poor sleep to long-term issues such as genetic mutations in offspring, irreversible crimes by intoxicated teenagers, and fatal traffic incidents from drunk driving—the number of alcohol consumers remains substantial.

More surprisingly, during the same year, 28 million people developed Alcohol Use Disorder (AUD), 4 million emergency department visits were alcohol-related, and about 178,307 deaths were linked to alcohol \cite{niaaa1}. 
According to \textit{National Institute on Alcohol Abuse and Alcoholism} \cite{gunzerath2004national}, alcohol adversely affects body organs including the brain, heart, liver, and pancreas, and it is also a known carcinogen. 
Alcohol interferes with the communication pathways of the brain and can influence the way the brain looks and works. 
These disruptions can also impact mood and behavior, impairing one's clarity of thought and ability to coordinate. 
In addition, excessive or acute heavy drinking can damage the heart, leading to stroke and high blood pressure.

The situation among underage populations regarding alcohol use is particularly alarming. According to the 2023 National Survey on Drug Use and Health (NSDUH), around 27.9\% (10.6 million) of teenagers aged 12 to 20 reported drinking in the past year, and 14.6\% (5.6 million) in the past month \cite{niaaa2}. 
Furthermore, 1.7\% (663,000) engaged in heavy alcohol use during the past month \cite{niaaa2}. 
Alcohol use in teenagers is reported to affect normal brain development and increase the risk of AUD. The teenage years are marked by significant growth and maturation in various body systems, including the brain. Alcohol use during this formative time can disrupt these developmental processes, potentially causing long-term changes in brain function and structure. This interference can heighten the likelihood of developing AUD and impact cognitive and behavioral functions over the long term. 
Given the challenges teenagers face with self-regulation and decision-making, factors such as family background and social interactions play significant roles in shaping their behaviors regarding alcohol use, which is crucial in preventing AUD among this demographic.

Alcohol Use Disorder is a chronic, relapsing brain disorder marked by an inability to regulate alcohol consumption, which can result in negative social, work-related, or health outcomes \cite{carvalho2019alcohol}. It is associated with various symptoms, including intense cravings for alcohol, difficulty controlling intake, increased tolerance (requiring more alcohol to achieve the same effect), and withdrawal symptoms when alcohol use is reduced or stopped. AUD is affected by a variety of genetic, environmental, psychological, and social factors. Understanding how each of these elements contributes to the development and progression of AUD is essential for developing effective prevention strategies. Therefore, it is important to examine the interaction between these factors and their impact on the onset of AUD.

In this study, we aimed to explore various factors, including individuals' background information, lifestyle choices, and family history of AUD. Our analysis utilized real-world survey data collected by the All of Us Research Program from the National Institutes of Health (NIH). We analyzed the data to examine how each factor influences alcohol consumption through classical machine learning techniques. Using decision trees, we ranked the importance of each attribute related to AUD, with higher importance indicating a stronger association with the disorder. This suggests that individuals can lower their risk of developing AUD by actively modifying specific aspects of their lifestyle and environment. Additionally, we applied several classical machine learning techniques, including decision trees, random forests, and Naive Bayes, to classify whether an individual has AUD based on their background and lifestyle information. These predictions can serve as an early intervention strategy to help prevent AUD.

\section{Objectives}


\subsection{Objectives}
As highlighted in the introduction, alcohol consumption, especially AUD, has profound negative effects on individuals and society. It is essential to identify the key factors contributing to alcohol use, particularly related to an individual’s background, lifestyle, and family history of AUD. These findings can support parents and teachers in taking preventive actions before children or students develop harmful drinking habits, which may lead to severe outcomes, such as dropping out of school. Similarly, adults can be encouraged to modify their lifestyle and habits to reduce the likelihood of developing AUD. The specific objectives of this work are outlined below.

\begin{itemize}
\item Identify the most influential determinants to reveal how lifestyle, background, and family history of AUD contribute to an individual's susceptibility to developing AUD. This can enable parents and educators to guide students with strategies to mitigate their AUD development. Similarly, adults may modify their behaviors or environments to minimize their own risk from AUD.
\item Predict the likelihood of an individual developing AUD by considering factors of lifestyle, background, and family history of AUD. This approach enables a proactive prediction of the individual’s risk of developing AUD, allowing for early intervention and targeted prevention.
\end{itemize}

\subsection{Data}\label{subsec: data}
The data used in this work is from the All of Us Research Program’s Registered Tier Dataset V7, available to authorized users on the Researcher Workbench. 
The All of Us Research Program, led by the NIH, aims to collect health data from over one million diverse participants across the U.S., with 409,420 enrolled so far. The program aims to explore how differences in biology, environment, and lifestyle impact health by creating a comprehensive health database. Participants contribute through surveys, electronic health records, physical measurements, and genetic samples in the dataset \cite{allofus_2024}.
There are three surveys included in the dataset we are using:
\begin{enumerate}
    \item The Basics: it gathers fundamental demographic information, including questions about participants' work and home life. Most of the questions and answers are extracted from the survey and added to our dataset. 
    \item Lifestyle: it surveys participants' use of tobacco, alcohol, and recreational drugs. There are also questions concerning user behaviors, such as usage frequency and duration of usage. Only questions related to participants' use of tobacco, alcohol, and recreational drugs are included in our dataset.
    \item Personal and Family Health History: it collects information about participants' past medical history, including their medical conditions, the approximate age at which they were diagnosed, and those of their family members. While the survey encompasses a wide range of diseases, our focus is solely on the information related to AUD for our research.
\end{enumerate}
The labels and brief description of the survey questions included in our dataset are shown in Table \ref{table: Allvar_brief}. The details can be found in the Appendix. We have 33 factors (i.e. features) in total. 

\begin{table}[ht]
\caption{The labels and brief descriptions of all the questions in our dataset, which are from one of the three surveys by the All of Us Program.}
\label{table: Allvar_brief}
\begin{center}
\begin{tabular}{|c|l|l|}
\hline
Survey & Questions & Brief description  \\
\hline
\hline
1 & BirthPlace & Country born in \\
\hline
1 & Race & Race Ethnicity \\
\hline 
1 & Gender & Gender identity \\
 \hline
1 & Sex & Biological Sex At Birth\\
\hline
1 & Education & Education Level: Highest Grade \\
\hline
1 & Marriage & Current marital status\\
\hline
1 & HealthInsurance & Health Insurance\\
\hline
1 & Hearing & Disability: Deaf\\
\hline
1 & Vision & Disability: Blind \\
\hline
1 & Concentration & Disability: Difficulty Concentrating \\
\hline
1 & WalkClimb & Disability: Walking Climbing \\
\hline
1 & DressBath & Disability: Dressing Bathing \\
\hline
1 & Errands &  Disability: Errands Alone \\
\hline
1 & Employment & Employment: Employment Status  \\
\hline
1 & AnnualIncome & Annual Income\\
\hline
1 & HomeOwn & Current Home Own \\
\hline
1 & LivingYear & Living years at the current address \\
\hline
1 & HouseholdNum & Living Situation: How Many People \\
\hline
1 & Household18Num & Living Situation: People Under 18 \\
\hline
1 & HouseholdStable & Stable House Concern \\
\hline
2 & Alcohol & Ever drink alcohol\\
\hline
2 & Cigar & Ever smoke a cigar\\
\hline
2 & ElecSmoke & Ever use an electronic nicotine product\\
\hline
2 & HookahSmoke & Ever smoke tobacco in a hookah\\
\hline
2 & RecreationDrug & Ever use recreational drugs\\
\hline
2 & TabaccoSmokeless & Ever use smokeless tobacco products\\
\hline
2 & SmokeQuit & A serious attempt to quit smoking\\
\hline
3 & AUD-Grandparent & Grandparent with AUD\\
\hline
3 & AUD-Mother & Mother with AUD\\
\hline
3 & AUD-Father & Father with AUD\\
\hline
3 & AUD-Sibling & Sibling with AUD\\
\hline
3 & AUD-Daughter & Daughter with AUD\\
\hline
3 & AUD-Son & Son with AUD\\
\hline
3 & AUD-Self & Self with AUD\\
\hline
\end{tabular}
\end{center}
\end{table}

\section{Methods}

\subsection{Data Preprocess} 
Each of the three surveys has more than 100,000 participants. We focused on participants who completed all three surveys in full. In other words, those with incomplete responses—including answers marked as ``Skip'', ``Prefer Not To Answer'', ``Don't Know'', or ``Response removed due to invalid value''—were excluded from our dataset. Altogether, the dataset includes 6,016 participants with complete information. Among the participants, 1,442 ($24\%$) have experienced AUD, while 4,574 ($76\%$) have never had AUD in their lifetime.

We found that the answers for sex at birth (Sex in Table~\ref{table: Allvar_brief}) and gender identity (Gender in Table~\ref{table: Allvar_brief}) are identical. In other words, all selected participants' personal sense of their gender aligns with the sex assigned to them at birth. 
If two features are highly correlated, the decision tree may choose only one for splitting at the nodes. After selecting the first feature, the second may not add any valuable information, leading to high importance for the first feature and low or zero importance for the second. Therefore, we excluded the gender data to avoid redundancy. 

Additionally, the question about alcohol (Alcohol in Table~\ref{table: Allvar_brief}) asks, "In your entire life, have you had at least one drink of any kind of alcohol?" Since $99.6\%$ of participants responded 'yes,' this provides negligible information for assessing alcohol use disorder (AUD). Therefore, we excluded this data.

In summary, a total of 31 features were included in our study for assessing and predicting AUD.


\subsection{Models}
\label{Models}
To explore the relationship between individual attributes and AUD, machine learning techniques are more appropriate for managing complex data than traditional statistical methods \cite{ebrahimi2022identification}. 

To understand the impact of lifestyle, background, and family history on the development of AUD (Objective 1), the importance of these factors (i.e., features) was analyzed.
 Machine learning techniques used for feature importance include: 1) tree-based models: decision trees \cite{kingsford2008decision}, random forests \cite{breiman2001random}, and gradient boosting \cite{friedman2001greedy}, 2) permutation feature importance \cite{altmann2010permutation}, 3) coefficients in linear models \cite{montgomery2021introduction, menard2002applied}, and 4) recursive feature elimination \cite{chen2007enhanced}. 
We chose decision trees to assess feature importance due to their ability to naturally rank features based on their contribution to the model \cite{kingsford2008decision}. 
After ranking the factors based on feature importance, we examined the top factors individually using data visualization to gain direct insights into their linear relationships with AUD. 
If the visualizations did not reveal clear relationships (e.g., between AUD-Grandparent and AUD-Self), we applied statistical methods as an alternative method. The Chi-Square Test of Independence \cite{mchugh2013chi} was selected as the alternative since factors in the dataset are represented as binary nominal variables.

After determining the importance of the features, the next step (Objective 2) was to predict an individual's likelihood of developing AUD by utilizing the identified features.
The classical machine learning techniques for classification include: 
1) Tree-based models: Tree-based models are easy to interpret and can handle mixed data types \cite{loh2011classification, james2013introduction}. However, small changes in the data can lead to different tree structures \cite{breiman2001random, friedman2001greedy}. 
2) Parametric models including linear regression \cite{montgomery2021introduction} and Naive Bayes \cite{rish2001empirical}: Parametric models are generally simpler and faster to train than non-parametric models, but they are less flexible in capturing complex relationships compared to tree-based models \cite{bishop2006pattern}. 
3) Instance-based learning algorithms including k-Nearest Neighbors (k-NN) \cite{cover1967nearest} and Radius Neighbors Classifier \cite{hartigan1979algorithm}: Instance-based learning algorithms are effective for modeling complex decision boundaries but are computationally intensive and sensitive to noise \cite{bishop2006pattern, cover1967nearest}. 

We chose the first two types of techniques for their computational efficiency. Specifically, we applied three classification techniques (Decision Trees, Random Forests, Naive Bayes) and compared their classification performance. Details of the selection can be found in the Appendix.

All assessments and predictions were conducted using Python 3.10.12 on the research workbench of the All of Us program platform. To ensure sufficient information for feature analysis, we used $80\%$ of the data for training and the remaining $20\%$ for testing. In addition to the standard train-test split (i.e., holdout validation \cite{lachenbruch1968estimation}, which splits the data once), we also perform classification using 5-fold cross-validation \cite{stone1974cross}, which splits the data into five subsets, reducing the risk of overfitting to a specific data split.
\section{Results and Discussion}
\label{Results}

\subsection{Importance of Each Attribute}
\begin{figure}[ht!]
  \centering
   \includegraphics[width=1\columnwidth]{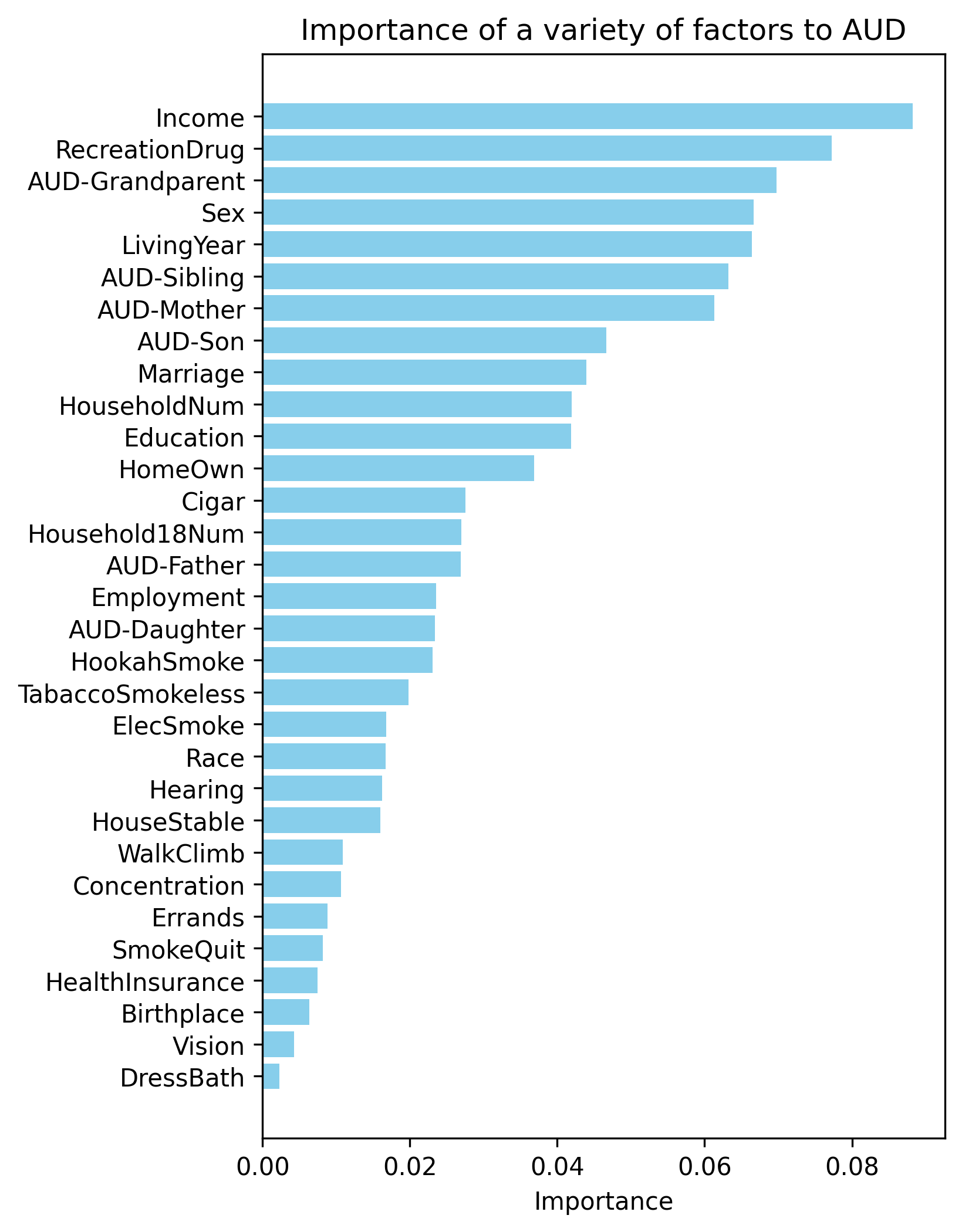}
  \caption{The importance of each factor from the three surveys (i.e., features) in relation to whether the individual has AUD.}
  \label{fig:importance}
\end{figure}

The feature importance of participants' lifestyle, background, and family history of AUD derived from Decision Trees is shown in Figure~\ref{fig:importance}. 
Our analysis reveals that the most significant factors (i.e. the ones with the highest feature importance) associated with AUD are individual annual income (0.0882), recreational drug use (0.0773), and the AUD status of the participant's grandparents (0.0698). 

Since the goal of this study is to identify modifiable or contextual factors that influence an individual's risk for AUD, and sex at birth is a fixed biological factor, we chose to focus more closely on the 10 other top factors. By examining these factors, such as family history and behavioral patterns, we can better identify specific areas where interventions or prevention strategies can be implemented.
These factors come from either \textit{The Basics} survey or  \textit{Personal and Family Health History} survey. 

We find that the factors from health history (i.e., AUD-Grandparent, AUD-Father, AUD-Mother, AUD-Sibling, AUD-Son, AUD-Daughter) do not exhibit a linear relationship with AUD from data visualization.
However, the importance of most of these features is high, such as AUD-Grandparent, AUD-Mother, AUD-Sibling, and AUD-Son. 
This may point to a non-linear relationship, where the effect of family history on AUD may not follow a simple, proportional pattern. 
Instead, it likely interacts with other factors, such as the individual's environment, upbringing, or coping mechanisms, which may amplify or mitigate the development of AUD. 

Alternatively, we use the Chi-Square Test of Independence to analyze the relationship between each of the AUD family factors and AUD (i.e. AUD-Self), and we find most of them display significant association as the p-values are smaller than 0.001 except AUD-Son and AUD-Daughter:
\begin{itemize}
    \item AUD-Grandparent: $\chi^2 = 38.86 $, $p = 4.54\times10^{-10}$
    \item AUD-Sibling: $\chi^2 = 30.81 $, $p = 2.85\times10^{-8}$
    \item AUD-Mother: $\chi^2 = 13.09 $, $p = 2.97\times10^{-4}$
    \item AUD-Father: $\chi^2 = 75.11 $, $p = 4.45\times10^{-18}$
\end{itemize}
The results indicate that AUD-Father has the strongest association with AUD-Self compared to the other variables. However, this only reflects a statistically stronger relationship and does not necessarily imply greater importance for classification. Similarly, although there is no significant association between AUD-Son/AUD-Daughter and AUD-Self, this does not negate their potential importance in classification.

\begin{figure}[ht!]
  \centering
   \includegraphics[width=0.9\columnwidth]{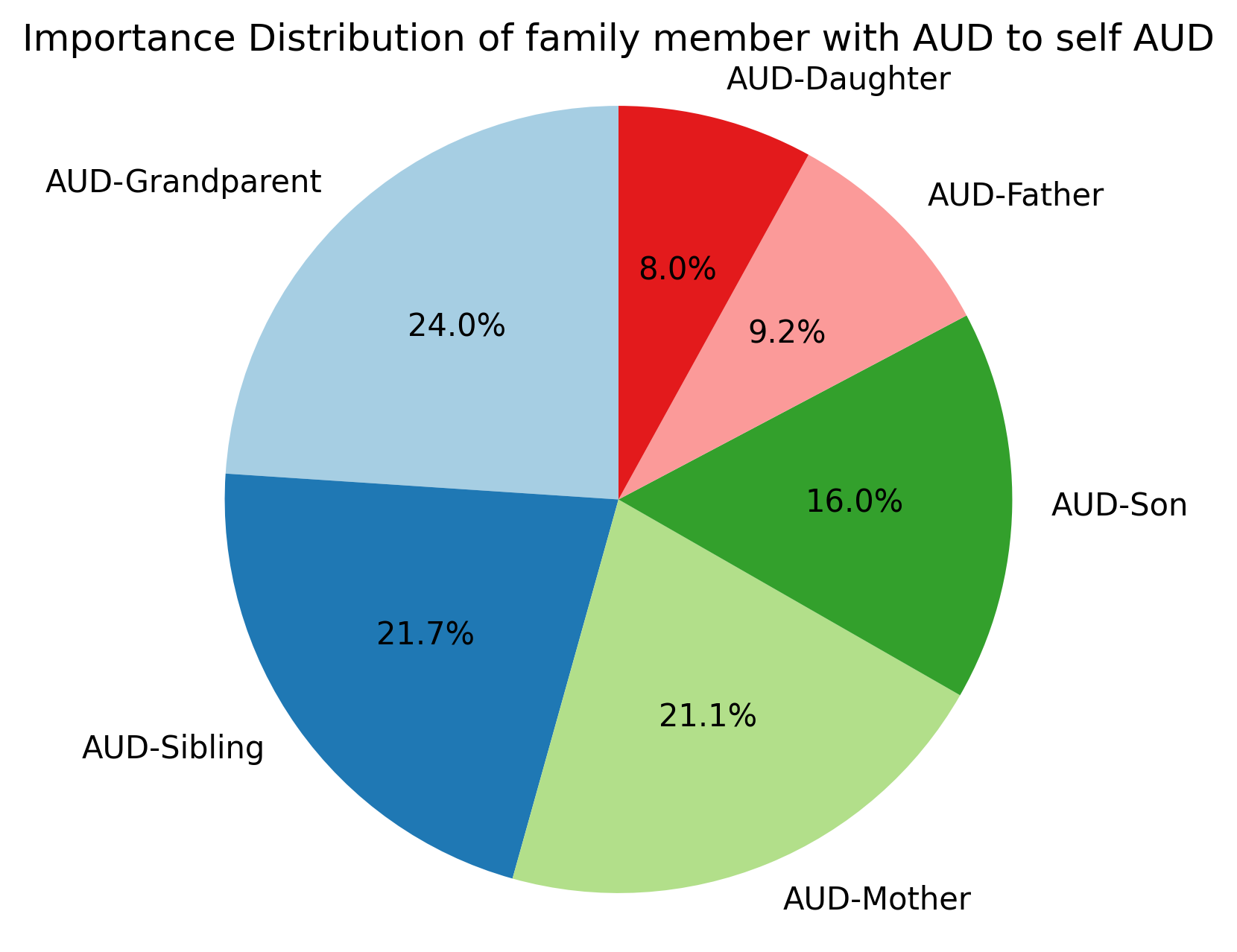}
  \caption{The importance of whether other family members have AUD in relation to the individual's AUD status.}
  \label{fig:famimportance}
\end{figure}

Figure~\ref{fig:famimportance} illustrates the importance of each of the six factors related to AUD. It indicates that the AUD status of a participant's grandparents, mother, and siblings has a stronger impact on their potential for developing AUD, compared to other factors. In addition to examining the family history of AUD across different family roles, we also analyzed it through a collective factor that considers whether any family member, other than the individual, has had AUD. The findings indicate even greater importance, with details provided in the Appendix.

Next, we find the other top factors from \textit{The Basics} Survey all exhibit a comparably strong linear relationship with AUD. Due to page constraints, the figures illustrating the relationships between these factors and AUD are included in the Appendix.
\begin{itemize}
    \item Annual Income: individuals with a lower annual income are more likely to develop AUD compared with others with a higher annual income. For example, there are $31.74\%$ who have had AUD among all the individuals with annual income less than $10,000$, while there are only $19.6\%$ who have had AUD among all the individuals with annual income more than $200,000$.
    \item Recreation drug use: individuals with no history of recreational drug use have the lowest likelihood of developing AUD compared to those who have used recreational drugs. Specifically, among individuals who do not use recreational drugs, $14.97\%$ have AUD, whereas $51.16\%$  of those who have used street opioids do have AUD.
    \item Living years: there are no significant differences in the AUD ratio (from $21.69\%$ to $26.15\%$) across different lengths of residence. However, participants who have lived at their current address for less than one year have the highest AUD ratio at $26.15\%$. It indicates a few possibilities: recent movers may face transitions that elevate stress levels, increasing AUD risk; new residents often lack established support networks, leading to potential isolation and heightened AUD risk; and recent movers might turn to alcohol as a coping strategy for the challenges of relocating.
    \item Sex: females are less likely to develop AUD compared with males. Specifically, $16.12\%$ of participants with AUD are female, while $38.21\%$ are male.
    \item Marriage status: individuals who are never married or separated have a higher likelihood of developing AUD compared to those who are married or widowed. Specifically, $20.87\%$ (the lowest) of participants with AUD are married, while $34.48\%$ (the highest) are separated. This may indicate that relationship stability and support can play a role in mitigating the risk of alcohol use disorder. 
    \item Education level: individuals with lower educational attainment, particularly those with a high school education or less, have a higher likelihood of developing AUD compared to those with a college degree or higher. Specifically, $22.27\%$ (the lowest) of participants with AUD are college graduates or hold advanced degrees, while $27.54\%$ (the highest) have completed high school or obtained a GED.
    This highlights the potential impact of education on alcohol use disorder risk. 
\end{itemize}

\subsection{Prediction: AUD or NonAUD}
The prediction results of the three different methods are shown in Table~\ref{table: accuracy} below and Table~\ref{table: prediction} in the Appendix. The prediction using Random Forests is the best among all three machine learning techniques. While the model achieves an overall accuracy of $81\%$ with regular validation, it performs well in identifying cases without AUD but struggles to accurately predict the cases with AUD. The main cause for this imbalance is likely due to the imbalance of the data (support: 284 with AUD, 920 without AUD). Addressing class imbalance and further optimizing the model could enhance its performance on the minority class. One effective approach is to use cross-validation, which led to improved performance for all three methods, achieving $82\%$ accuracy for random forests. However, since the improvement is insufficient, future work can explore additional methods such as resampling and stratified cross-validation.

\begin{table}[ht!]
\caption{Classification Accuracy by using three different machine learning (ML) techniques. Accuracy-1 fold refers to the accuracy obtained using holdout validation, while Accuracy-5 fold refers to the accuracy from 5-fold cross-validation.}
\label{table: accuracy}
\begin{center}
\begin{tabular}{|c|c|c|c|}
\hline
ML Technique & Decision Trees & Random Forests & Naive Bayes \\
\hline
Accuracy-1 fold & 0.71 & 0.81 & 0.70 \\
\hline
Accuracy-5 fold& 0.74 & 0.82 & 0.71 \\
\hline
\end{tabular}
\end{center}
\end{table}

\section{Conclusion}
By using Decision Trees, we first identified the most important features influencing alcohol use disorder (AUD): annual income, recreational drug use, length of residence, sex/gender, marital status, education level, and family history of AUD. This information can be valuable for parents, healthcare professionals, and educators in guiding individuals to reduce the risk of developing AUD before it becomes a serious issue. We then utilized data visualization and Chi-Square Tests of Independence to analyze the associations between individual factors and AUD, aiming to enhance our understanding of these relationships. Last but not least, we applied three different machine learning techniques to predict whether an individual has developed AUD based on 31 factors from lifestyle, personal information, and family history of AUD. Among these methods, random forests achieved the highest accuracy ($81\%$), compared to Decision Trees and Naive Bayes.

\section*{Acknowledgement}
We gratefully acknowledge All of Us participants for their contributions, without whom this research would not have been possible. We also thank the National Institutes of Health’s All of Us Research Program for making available the participant data examined in this study.


\nocite{*}
\bibliographystyle{unsrt}
\bibliography{ref}

\newpage
\section{Appendix}
\subsection{The detailed questions and the answers from the surveys of the All of Us program. }
Answers marked as “Skip”,
“Prefer Not To Answer”, “Don’t Know”, or “Response removed due to invalid value” are excluded.

(1) \textit{The Basics} survey:

\begin{table}[ht!]
\caption{The questions and the answers from \textit{The Basics} survey.}
\label{table:basic-var}
\begin{center}
\begin{tabular}{|p{0.21\linewidth}|p{0.69\linewidth}|}
\hline
\hline
Attribute & Question (Q) and Answers (A) \\
\hline
\hline
BirthPlace & Q: In what country were you born? \\
 & A: USA or Other (binary)\\
\hline
Race & Q: Which categories describe you? Select all that apply. Note, you may select more than one group. \\
  & A: White, Black, Hispanic, Asian, More than one, Other, Race Ethnicity None Of These (nominal)\\
\hline 
Gender & Q: What terms best express how you describe your gender identity? (Check all that apply) \\
 & A: Man, Woman (binary)\\
 \hline
Sex & Q: What was your biological sex assigned at birth? \\
& A: Female or Male (binary)\\
\hline
Education &Q: What is the highest grade or year of school you completed? \\
& A: 1-11 or never attended, 12 or  GED, College 1-3, College graduate or Advanced degree (nominal)\\
\hline
Marriage & Q: What is your current marital status? \\
& A: Married, Never married, Divorced, Living with partner, Widowed, Separated (nominal)\\
\hline
HealthInsurance & Q: Are you covered by health insurance or some other kind of health care plan? \\
& A: Yes or No (binary)\\
\hline
Hearing & Q: Are you deaf, or do you have serious difficulty hearing? \\
& A: Yes or No (binary)\\
\hline
Vision & Q: Are you blind, or do you have serious difficulty seeing, even when wearing glasses? \\
& A: Yes or No (binary)\\
\hline
Concentration & Q: Because of a physical, mental, or emotional condition, do you have serious difficulty concentrating, remembering, or making decisions? \\
& A: Yes or No (binary)\\
\hline
WalkClimb & Q: Do you have serious difficulty walking or climbing stairs?\\
& A: Yes or No (binary)\\
\hline
DressBath & Q: Do you have difficulty dressing or bathing?\\
& A: Yes or No (binary)\\
\hline
Errands & Q: Because of a physical, mental, or emotional condition, do you have difficulty doing errands alone such as visiting a physician's office or shopping? \\
& A: Yes or No (binary)\\
\hline
Employment & Q: What is your current employment status? Please select 1 or more of these categories. \\
& A: Employed for wage, Retired, Unable to work, Self-employed, Out of work one or more, Student, Homemaker, Out of work less than one (nominal)\\
\hline
AnnualIncome & Q: What is your annual household income from all sources? \\
& A: $< 10k$, $10k-25k$, $25k-35k$, $35k-50k$, $50k-75k$, $75k - 100k$, $100k- 150k$, $150k - 200k$, $> 200k$ (numeric)\\
\hline
HomeOwn & Q: Do you own or rent the place where you live? \\
& A: Own or Rent (binary)\\
\hline
HouseholdNum & Q: How many people live in the same house with you? \\
& A: 0,1,2,3,4,5,6,7,8,9,10,11 or more (numerical)\\
\hline
Household18Num & Q: How many people under 18 live with you? \\
& A: 0,1,2,3,4,5,6 or more (numerical)\\
\hline
\hline
\end{tabular}
\end{center}
\end{table}

\begin{table}[ht!]
\caption{The questions and the answers (continuing) from \textit{The Basics} survey.}
\label{table:basic-var2}
\begin{center}
\begin{tabular}{|p{0.15\linewidth}|p{0.75\linewidth}|}
\hline
\hline
Attribute & Question (Q) and Answers (A) \\
\hline
\hline
HouseStable & Q: Do you have any concern about the stability of your house? \\
& A: Yes or No (binary)\\
\hline
LivingYear & Q: How many years have you lived at your current address? \\
& A: $<1$, $1-2$, $3-5$, $6-10$, $11-20$, $>20$ (numeric)\\
\hline
\end{tabular}
\end{center}
\end{table}

(2) \textit{Lifestyle} survey:
\begin{table}[ht!]
\caption{The questions and the answers from \textit{The Basics} survey.}
\label{table:lifestyle-var}
\begin{center}
\begin{tabular}{|p{0.25\linewidth}|p{0.65\linewidth}|}
\hline
\hline
Attribute & Question (Q) and Answers (A) \\
\hline
\hline
Alcohol & Q: In your entire life, have you had at least 1 drink of any kind of alcohol, not counting small tastes or sips? (By a “drink,” we mean a can or bottle of beer, a glass of wine or a wine cooler, a shot of liquor, or a mixed drink with liquor in it.)  \\
 & A: Yes or No (binary)\\
 \hline
Cigar & Q: Have you ever smoked a traditional cigar, cigarillo, or filtered cigar, even one of two puffs? \\
 & A: Yes or No (binary)\\
  \hline
RecreationDrug & Q: In your LIFETIME, which of the following substances have you ever used? \\
 & A: Yes or No (binary)\\
   \hline
ElecSmoke & Q: Have you ever used an electronic nicotine product, even one or two times? (Electronic nicotine products include e-cigarettes, vape pens, hookah pens, personal vaporizers and mods, e-cigars, e-pipes, and e-hookahs.)\\
 & A: Cocaine', 'Hallucinogens', 'Inhalants', 'Marijuana', 'Methamphetamine', 'None', 'Others',  'Prescription Opioids', 'Prescription Stimulants', 'Sedatives', 'Street Opioids' (nominal)\\
\hline
TabaccoSmokeless & Q: Have you ever used smokeless tobacco products, even one or two times? (Smokeless tobacco products include snus pouches, Skoal Bandits, loose snus, moist snuff, dip, spit, and chewing tobacco.) \\
 & A: Yes or No (binary)\\
 \hline
SmokeQuit & Q: In the past, have you ever made a serious attempt to quit smoking? That is, have you stopped smoking for at least one day or longer because you were trying to quit? \\
 & A: Yes or No (binary)\\
\hline
\end{tabular}
\end{center}
\end{table}

(3) \textit{Personal and Family Health History} survey:

There is only one question about AUD in this survey: "Including yourself, who in your family has had alcohol use disorder? Select all that apply.". There are seven answers to choose from: Daughter, Father, Grandparent, Mother, Self, Sibling, and Son. Equivalently, we can break it into seven questions and each with a binary answer.
\begin{table}[ht!]
\caption{The questions and the answers from \textit{Personal and Family Health History} survey.}
\label{table:family-var2}
\begin{center}
\begin{tabular}{|p{0.2\linewidth}|p{0.7\linewidth}|}
\hline
\hline
Attribute & Question (Q) and Answers (A) \\
\hline
\hline
AUD-Family & Q: Has any family member, excluding yourself, had alcohol use disorder? \\
 & A: Yes or No (binary)\\
\hline
\end{tabular}
\end{center}
\end{table}

\begin{table}[ht!]
\caption{The questions and the answers from \textit{Personal and Family Health History} survey.}
\label{table:family-var}
\begin{center}
\begin{tabular}{|p{0.2\linewidth}|p{0.7\linewidth}|}
\hline
\hline
Attribute & Question (Q) and Answers (A) \\
\hline
\hline
AUD-Grandparent & Q: Has your grandparent had alcohol use disorder? \\
 & A: Yes or No (binary)\\
 \hline
AUD-Mother & Q: Has your mother had alcohol use disorder? \\
 & A: Yes or No (binary)\\
 \hline
AUD-Father & Q: Has your father had alcohol use disorder? \\
 & A: Yes or No (binary)\\
 \hline
AUD-Sibling & Q: Has your sibling had alcohol use disorder? \\
 & A: Yes or No (binary)\\
 \hline
AUD-Son & Q: Has your son had alcohol use disorder? \\
 & A: Yes or No (binary)\\
  \hline
AUD-Daughter & Q: Has your daughter had alcohol use disorder? \\
 & A: Yes or No (binary)\\
 \hline
AUD-Self & Q: Have you had alcohol use disorder? \\
 & A: Yes or No (binary)\\
\hline
\end{tabular}
\end{center}
\end{table}

An alternative approach to this question is to investigate whether any family member, aside from the individual, has had AUD (Table~\ref{table:family-var2}).

\section{Model details}
(1) Decision Trees: it is a supervised learning algorithm used for both classification and regression tasks. They operate by recursively partitioning the data into subsets based on feature values, forming a tree-like structure. In this structure, internal nodes represent decisions based on specific features, branches indicate the outcomes of those decisions, and leaf nodes represent the final predictions or outcomes. Due to the presence of 31 features, the generated decision tree is extensive and challenging to display here. However, it is possible to visualize the tree for detailed analysis.

(2) Random Forests: this technique constructs multiple decision trees during training and outputs the mode of their predictions. In addition to the advantages of decision trees, random forests offer several benefits, including a reduced risk of overfitting and improved predictive accuracy by averaging the results of multiple trees. They are also less sensitive to noise in the data compared to single decision trees and can be more computationally efficient due to parallel processing.

(3) Naive Bayes: this family of probabilistic algorithms is based on Bayes' theorem and assumes that the features used for classification are independent of each other. This assumption simplifies the computation of probabilities, resulting in efficient performance. Specifically, we chose Gaussian Naive Bayes because our data contains categorical features.

\section{Result and Discussion}
\subsection{Importance of Each Attribute}
The details of the linear relation between the top factors from \textit{The Basics} Survey and AUD can be found in  Figure 3 - Figure 8.


 \begin{figure}[ht!]
  \centering
   \includegraphics[width=0.8\columnwidth]{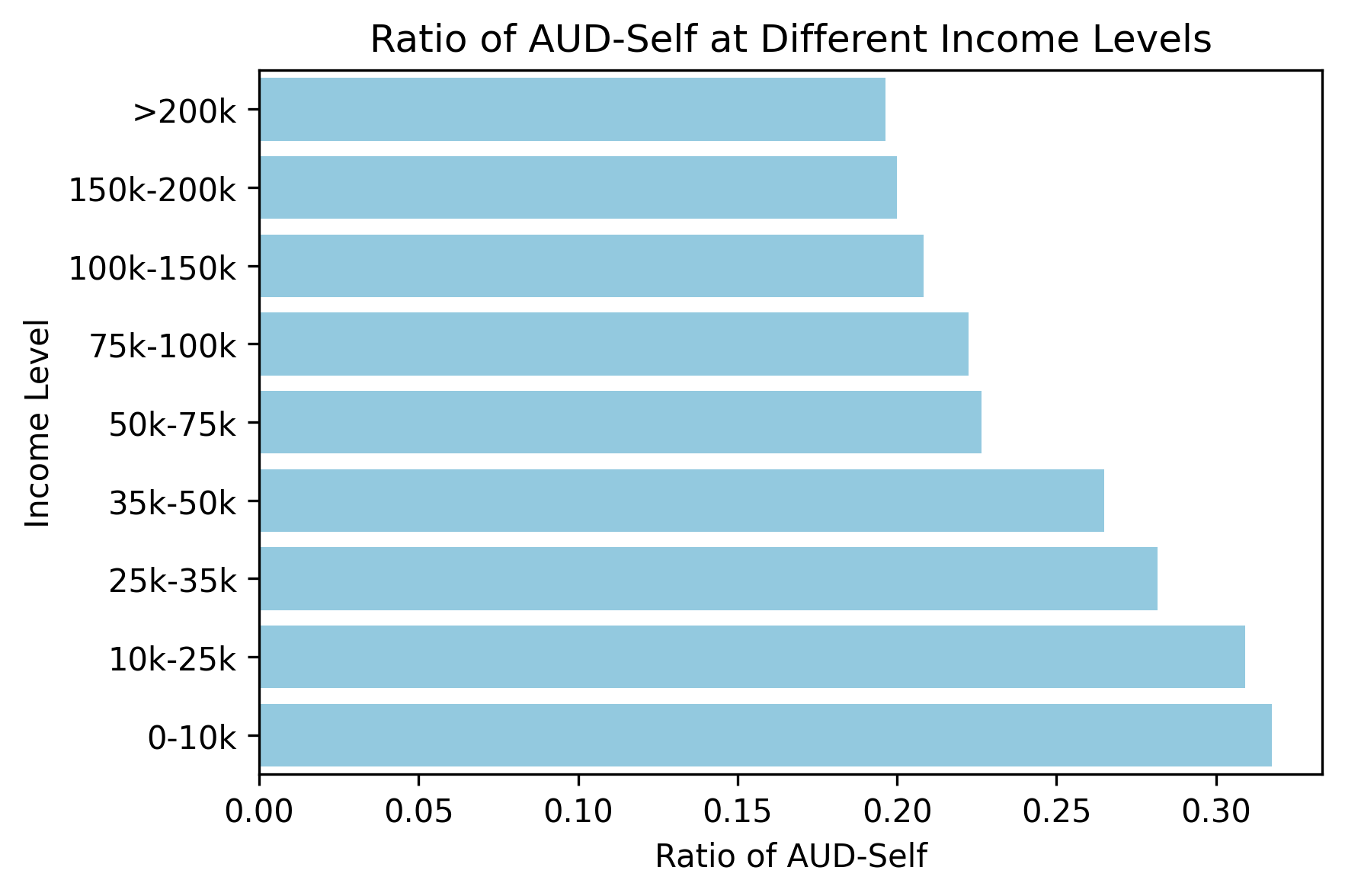}
  \caption{At each annual income level, the ratio of participants with AUD to the total number of participants at that income level is displayed.}
  \label{fig:income}
\end{figure}

    \begin{figure}[ht!]
      \centering   \includegraphics[width=0.8\columnwidth]{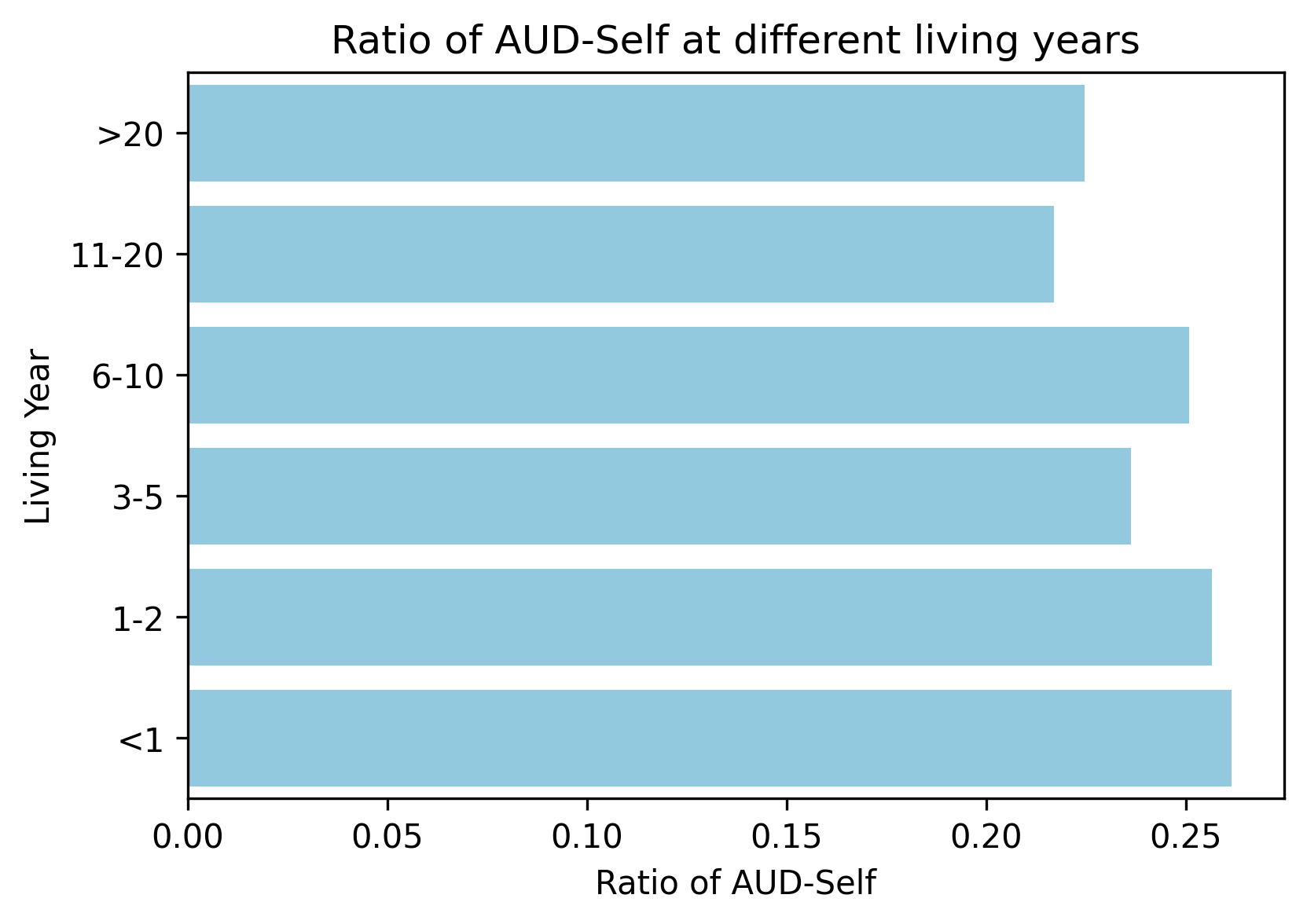}
      \caption{Given the duration of residence at the current address, the ratio of participants with AUD to the total number of participants sharing the same length of residence.}
      \label{fig:livingyear}
    \end{figure}
    
        \begin{figure}[ht!]
      \centering   \includegraphics[width=1\columnwidth]{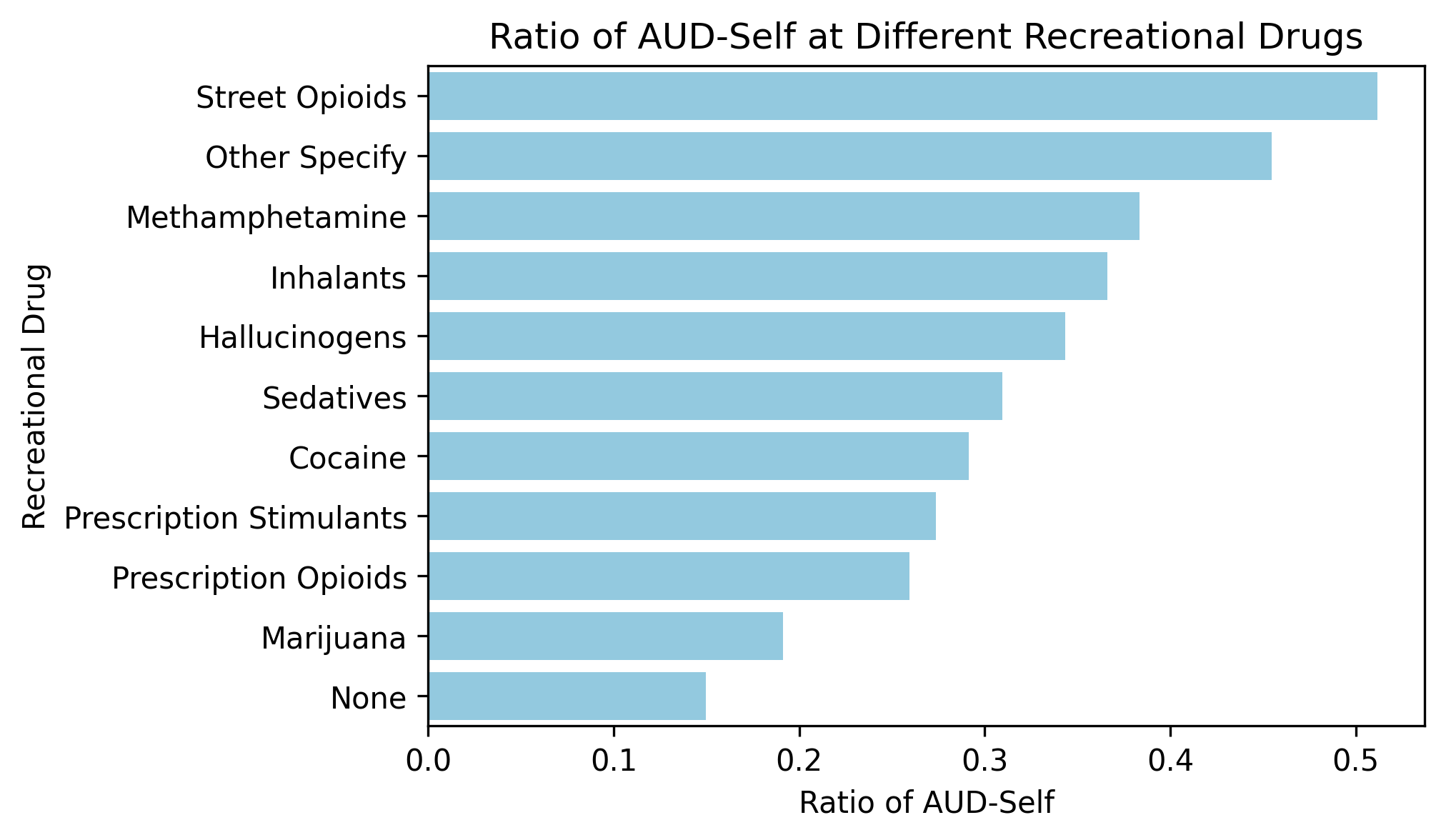}
      \caption{For each type of recreational drug, the ratio of participants with AUD to the total number of participants who have used that drug is presented.}
      \label{fig:drug}
    \end{figure}
        \begin{figure}[ht!]
      \centering   
      \includegraphics[width=0.8\columnwidth]{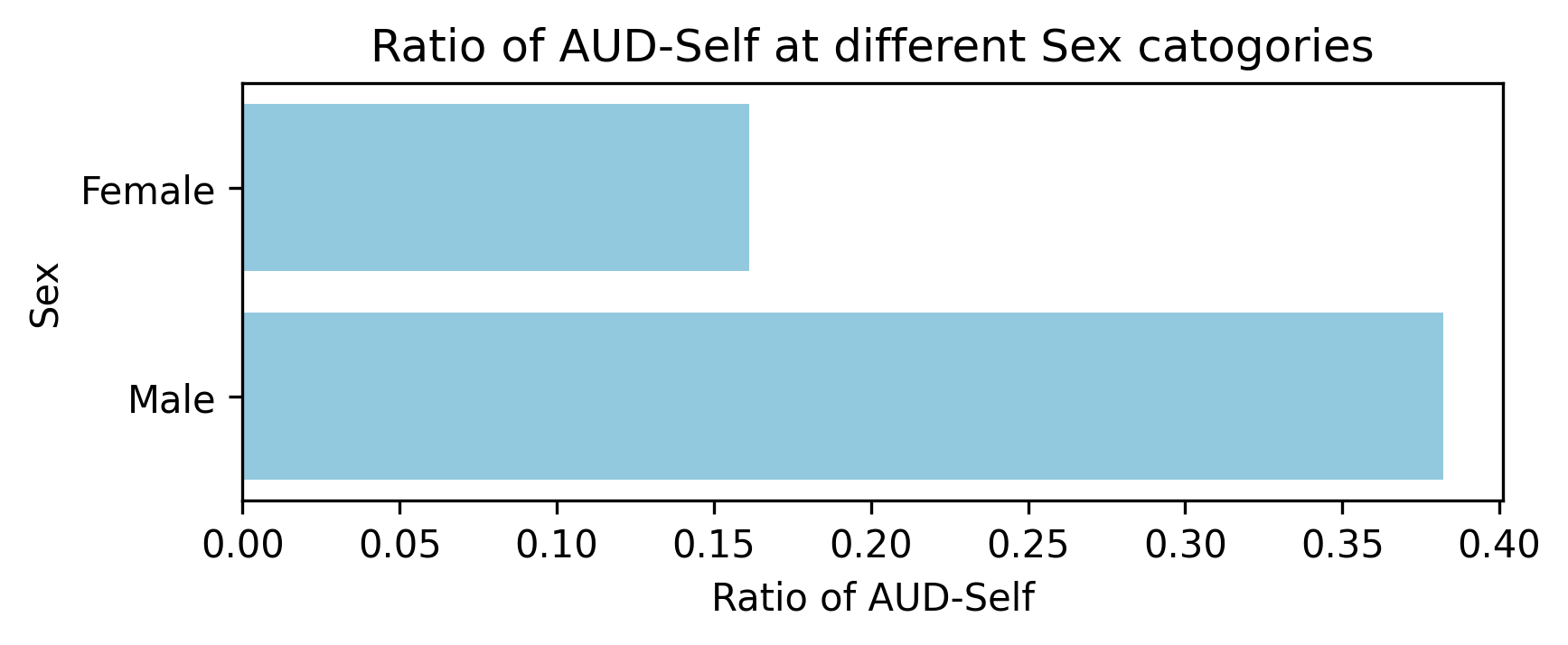}
      \caption{The ratio of participants with alcohol use disorder to the total number of participants, categorized by sex at birth, is presented.}
      \label{fig:SexHisRatio}
    \end{figure}

        \begin{figure}[h!]
        \centering   
        \includegraphics[width=0.8\columnwidth]{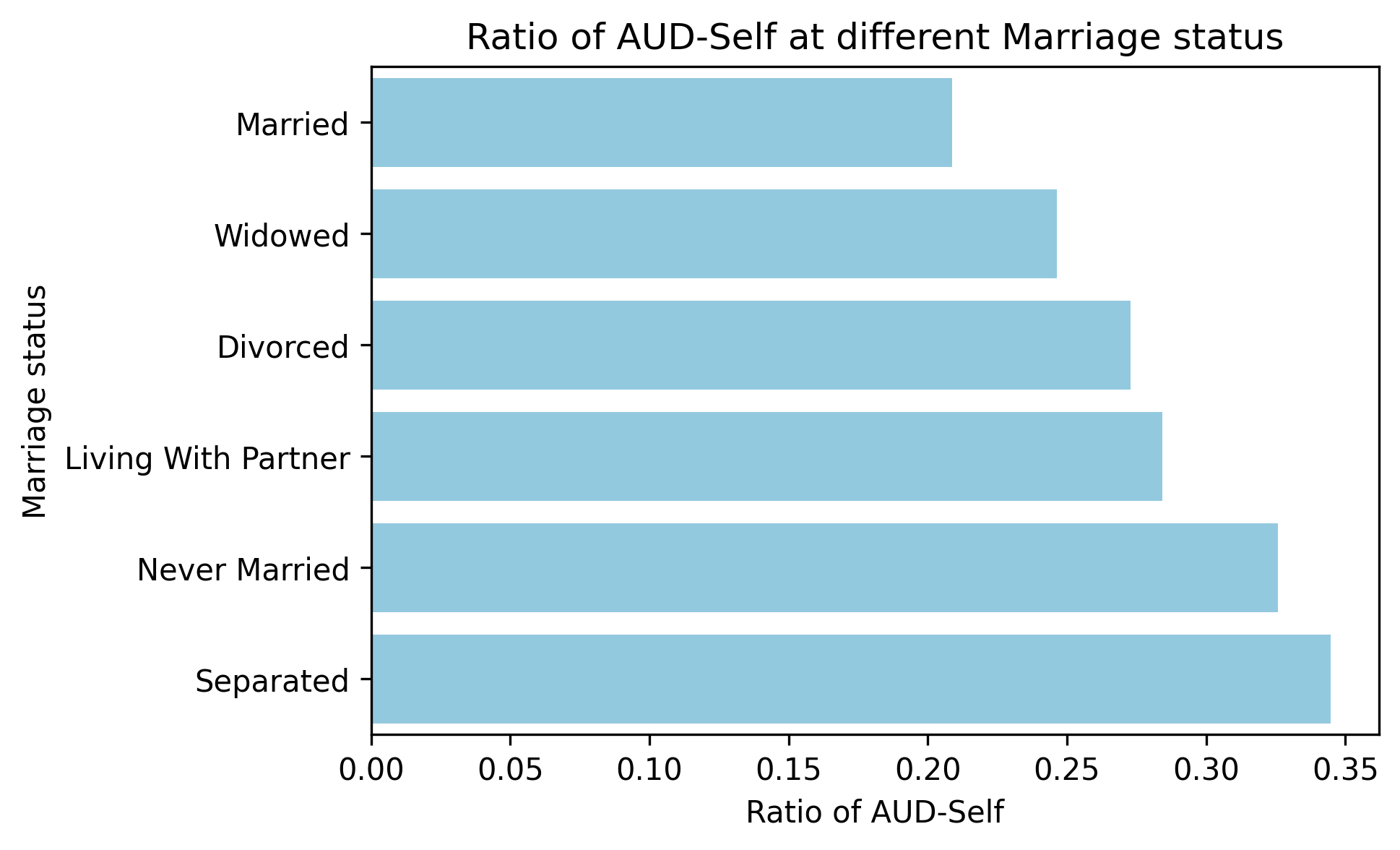}
        \caption{The ratio of participants with alcohol use disorder to the total number of participants, categorized by marriage status, is presented.}
        \label{fig:marry}
        \end{figure}
        \begin{figure}[ht!]
        \centering   
    \includegraphics[width=1\columnwidth]{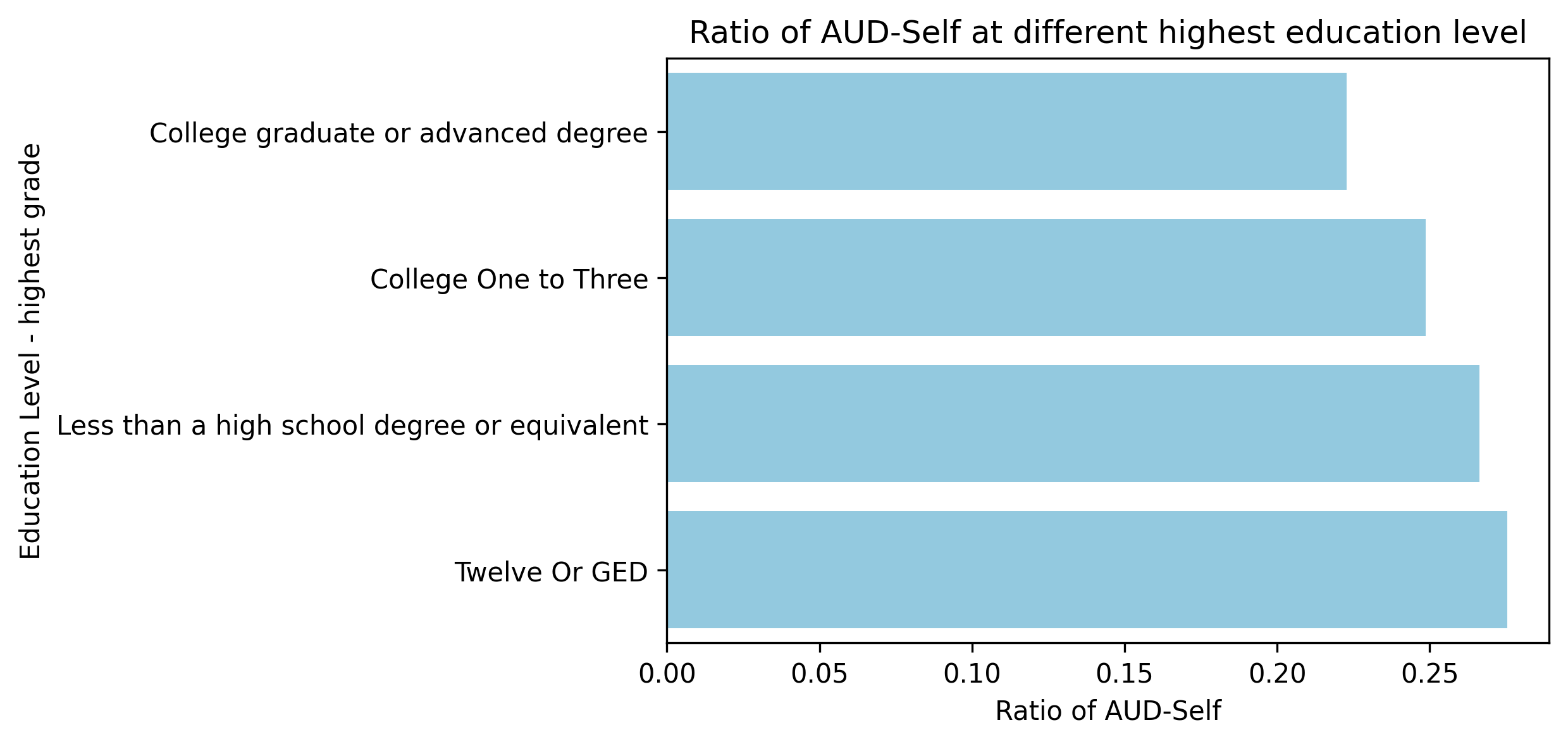}
        \caption{The ratio of participants with alcohol use disorder to the total number of participants is presented, categorized by highest education level.}
        \label{fig:edu}
         \end{figure}
         
\subsection{Prediction: AUD or NonAUD}
Table~\ref{table: prediction} shows the performance metrics of all three ML techniques for the AUD classification.
\begin{table}[ht!]
\caption{Classification performance metrics of three different ML techniques using holdout validation.}
\label{table: prediction}
\begin{center}
\begin{tabular}{|c|c|c|c|c|}
\hline
ML Technique &AUD-Self & precision & Recall & $f_1$ -score  \\
\hline
Decision Trees & no & 0.83 & 0.79 & 0.81 \\
                & yes & 0.41 & 0.48 & 0.44 \\
\hline
Random Forests & no & 0.82 & 0.97 & 0.89 \\
                & yes & 0.75 & 0.31 & 0.44 \\
\hline
Naive Bayes & no & 0.81 & 0.79 & 0.80\\
                & yes & 0.37 & 0.41 & 0.39 \\
\hline
\end{tabular}
\end{center}
\end{table}


\section{Results when combining all family AUD factors into one factor AUD-Family }
\subsection{Feature Importance}
\begin{figure}[ht!]
  \centering
   \includegraphics[width= 0.9\columnwidth]{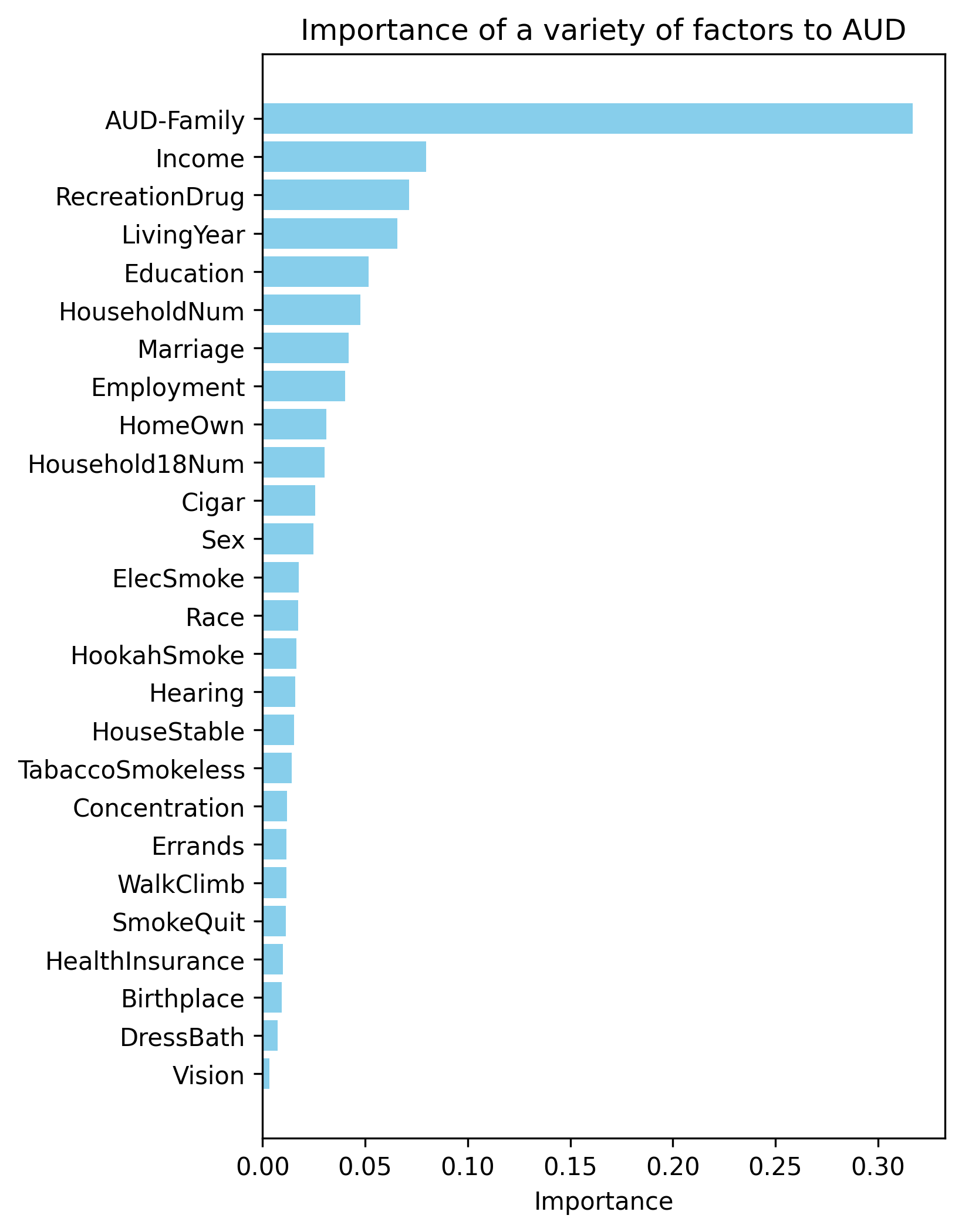}
  \caption{The importance of each factor from the three surveys (i.e., features) in relation to whether the individual has AUD.}
  \label{fig:importance2}
\end{figure}

$91.09\%$ of participants have one or more relatives with AUD in their close family. As shown in Figure~\ref{fig:importance2}, AUD-Family is the most significant factor, with a value of 0.3169, in predicting whether an individual has developed AUD. The feature importance rankings for other factors closely align with those derived when considering all seven AUD factors as features.
Besides, the Chi-Square Test of Independence results indicate a significant association between AUD-Family and the likelihood of developing AUD ($\chi^2 = 1852.76$, $p = 0.0000$).

\subsection{Prediction: AUD or NonAUD}
The prediction performances of all three are improved compared with our previous results. Random forests is still the best among them with high accuracy. The overall performance of the model is mixed, with strong results overall but notable room for improvement in identifying the minority class. Similarly, cross-validation led to improved performance for all three methods.
\begin{table}[ht!]
\caption{Classification Accuracy by using three different ML techniques with holdout and 5-fold cross-validation. }
\label{table: accuracy2}
\begin{center}
\begin{tabular}{|c|c|c|c|}
\hline
ML Technique & Decision Trees & Random Forests & Naive Bayes \\
\hline
Accuracy-1 fold & 0.749 & 0.836 & 0.828 \\
\hline
Accuracy-5 fold & 0.754 & 0.838 & 0.833  \\
\hline
\end{tabular}
\end{center}
\end{table}

\begin{table}[ht!]
\caption{Classification performance metrics of three different ML techniques.}
\label{table: prediction2}
\begin{center}
\begin{tabular}{|c|c|c|c|c|}
\hline
ML Technique &AUD-Self & precision & Recall & $f_1$ -score  \\
\hline
Decision Trees & no & 0.85 & 0.82 & 0.83 \\
                & yes & 0.47 & 0.52 & 0.50 \\
\hline
Random Forests & no & 0.84 & 0.98 & 0.90 \\
                & yes & 0.84 & 0.38 & 0.52 \\
\hline
Naive Bayes & no & 0.84 & 0.95 & 0.89\\
                & yes & 0.74 & 0.42 & 0.53 \\
\hline
\end{tabular}
\end{center}
\end{table}
\end{document}